\pdfoutput=1

\documentclass[11pt]{article}

\usepackage[final]{acl}

\usepackage{times}
\usepackage{latexsym}

\usepackage[T1]{fontenc}

\usepackage[utf8]{inputenc}

\usepackage{microtype}

\usepackage{inconsolata}

\usepackage{graphicx}
\usepackage{amsmath}
\usepackage{booktabs}
\usepackage{multirow}
\usepackage{makecell}
\usepackage{hyperref}
\usepackage{svg}
\usepackage{xurl}
\usepackage{bm}
\usepackage{enumitem}

\title{Factuality on Demand: Controlling the Factuality–Informativeness Trade-off in Text Generation}

\author{
  \textbf{Ziwei Gong$^\dagger$\textsuperscript{1}}, 
  \textbf{Yanda Chen$^\dagger$\textsuperscript{1}\thanks{Work done during PhD at Columbia University.}}, \\
   \vspace{3pt}
    \textbf{Julia Hirschberg\textsuperscript{1}}, 
    \textbf{Chen Zhao\textsuperscript{2,3}}, 
\textbf{He He\textsuperscript{2}},
\textbf{Zhou Yu\textsuperscript{1}}, 
  \textbf{Kathleen Mckeown\textsuperscript{1}}
\\
  \textsuperscript{1}Columbia University, 
  \textsuperscript{2}New York University,
\textsuperscript{3}NYU Shanghai,
\\
  \small{$^\dagger$Equal contributions.}\\
  \small{
    \texttt{ \{sara.ziweigong, yanda.chen\}@cs.columbia.edu }
  }
}

\begin{document}
\maketitle


\begin{abstract}

Large language models (LLMs) encode knowledge with varying degrees of confidence.
When responding to queries, models face an inherent trade-off: they can generate responses that are less informative but highly factual, or more informative but potentially less accurate.
Different applications demand different balances between informativeness and factuality.
We introduce Factuality-Controlled Generation (FCG), a framework that enables users to specify factuality constraints alongside their queries.
We propose to evaluate FCG performance on two dimensions: adherence to factuality constraints and response informativeness.
We propose to train models on the FCG task using synthetic data, and show that our synthetic training significantly improves models' ability to both respect factuality requirements and maintain informativeness in their outputs.

\end{abstract}

\section{Introduction}

Large language models (LLMs) are widely used for question answering, summarization, and content generation. Their internal knowledge, however, is unevenly reliable: some statements are strongly supported, while others are speculative, outdated, or uncertain. Generation therefore requires deciding how much to say and how cautiously to say it, creating a tension between factual precision and informativeness. Humans make analogous choices: starting with high-reliability facts and adding lower-certainty details only when asked \citep{bloomfield1996communication, yaniv1995graininess, schustek2019human}.

This trade-off is application-dependent. High-assurance settings (e.g., medical or legal) demand conservative, highly factual outputs; creative writing, tutoring, or brainstorming often prefer richer content despite some factual risk \citep{wang-etal-2024-factuality, wang2023survey, wei2024measuringshortformfactualitylarge}. However, current LLMs offer no built-in mechanism to control this trade-off. While users may try to guide the model’s behavior with prompts like “be more factual,” we find that frontier models do not reliably adjust their outputs in response to such prompts on this task.
On FactScore \citep{min-etal-2023-factscore}, we find off-the-shelf models often fail to satisfy even moderate-to-strict targets. This gap motivates a controllable alternative that lets users request a specific factuality level and have the model adjust its responses accordingly.

We propose Factuality-Controlled Generation (FCG), a task and framework for conditioning LLM outputs on a user-specified factuality level, expressed as a numeric threshold (e.g., 80\% of the information being correct). A response meets the constraint if the fraction of supported factual statements exceeds this level. We frame FCG as a controllable generation task over knowledge-intensive question answering, where the goal is to generate an informative yet reliable answer based on a user-specified factuality level.

We evaluate FCG along two axes that reflect the task’s goals: (1) factuality adherence, the proportion of generations that satisfy the requested factuality level, measured across the dataset; and (2) informativeness, the amount of supported content in the output, measured as the number of validated atomic statements, normalized by output length.

We evaluate frontier models on FCG with prompting and find that they fail to reliably respect factuality constraints.
A natural way to improve models' ability on FCG is through supervised training as a controlled generation task: given a query and a numeric factuality level, the model is trained to generate answers that meet the factuality constraint while remaining informative.
Unfortunately, there does not exist such labeled data of question-factuality pairs answered with with gold responses that respect the factuality constraint.
Therefore, we propose to train FCG on synthetic data.
To generate training data, we introduce a confidence-guided synthetic pipeline. For each question, we first elicit a detailed, unconstrained answer. We then score the confidence of its atomic statements, and remove the minimal subset of low-confidence content necessary to meet the desired factuality level. 
Repeating this process across a range of factuality levels (e.g., 0.8 to 1.0) yields aligned answer sets at multiple factuality levels, which we use to fine-tune the model to condition on the target at inference.
We fine-tune a Mistral-7B model on synthetic data we constructed, enabling it to dynamically adjust its responses based on the desired reliability level.

FCG synthetic training substantially improves adherence to factuality constraints compared to prompting baselines. 
The proportion of generations satisfying factuality levels of 0.8 and 0.9 increases by 17.2\% and 130.0\% respectively (relative gains); at the strict level of 1.0 (all information in the response being correct), the adherence rate improves from 0\% to 23.6\%. 
Importantly, these improvements do not come at the cost of informativeness: for matched factuality levels, FCG yields more supported information per response, shifting the factuality–informativeness trade-off frontier outward. 
In summary, FCG enables users to obtain better factuality-informativeness trade-offs with more reliable customization control that suits their individual use cases.

\section{Related Work}
\paragraph{Factuality}
\textit{Ensuring factual accuracy} in LLMs has become a critical area of research, as these models often generate information that appears plausible but is incorrect, a phenomenon commonly referred to as hallucination \cite{wang2023survey, wei2024measuringshortformfactualitylarge}.
Proposed strategies include integrating external knowledge sources \cite{tonmoy2024comprehensivesurveyhallucinationmitigation}, implementing self-reflection mechanisms \cite{gao2023retrieval}, and refining model outputs during the generation process \cite{rafailov2024direct}. These efforts aim to align LLM outputs with verified information, thereby reducing the incidence of false or unsupported statements. 
\citet{mohri2024language} propose the conformal factuality framework, which ensures high-probability correctness for language model outputs using conformal prediction without extensive human annotations. 
However, their method relies on post-hoc filtering during inference, which can be computationally expensive and may reduce response fluency. Unlike their approach, we fine-tune language models to internalize the factuality-informativeness trade-off during training. This enables the model to generate responses that meet user-specified factuality levels directly at inference time without requiring re-ranking or rejection sampling, thereby improving both efficiency and controllability.

\textit{Evaluating  factuality} is crucial for identifying potential errors and advancing the development of more reliable language models, particularly in open-ended text generation \cite{wang-etal-2024-factuality}. To address this, researchers have developed various benchmarks \cite{vu2023freshllmsrefreshinglargelanguage, lin-etal-2022-truthfulqa, hendryckstest2021} and evaluation methods \cite{wei2024measuringshortformfactualitylarge, min-etal-2023-factscore, manakul-etal-2023-selfcheckgpt, wang2023evaluating} to quantify factual correctness in model outputs. We build on this line of work by using an automatic factuality metric (FactScore) both to construct training data and to assess our model’s performance.

\paragraph{Informativeness}

Recent research has focused on evaluating and enhancing informativeness in language models. \citet{kamalloo2023hagridhumanllmcollaborativedataset}introduced HAGRID, a dataset combining human- and LLM-generated data to improve generative information-seeking models with attributed explanations. \cite{valentini-etal-2025-measuring} tackled contextual informativeness in child-directed texts by using LLMs to evaluate how well stories convey target vocabulary semantics, advancing educational content generation. \cite{liang2024antevalevaluationsocialinteraction} proposed AntEval, a framework for assessing social interaction competencies in LLM-driven agents, introducing metrics like Information Exchanging Precision (IEP) to measure interaction quality. \cite{fan2024evascoreevaluatingabstractivelongform} developed EVA-Score, an evaluation metric for abstractive long-form summarization that addresses the limitations of traditional metrics by focusing on informativeness and overlap with reference texts. Our work extends these efforts by introducing Factuality-Controlled Generation, which uniquely enables users to dynamically balance factuality and informativeness in the output, offering greater flexibility depending on the application’s requirements.

\paragraph{LLM Fine-tuning via Hindsight}
Hindsight-based methods such as RvS \cite{emmons2022rvs}, HER \cite{NIPS2017_453fadbd}, and HIR \cite{10.5555/3618408.3620145} improve learning by turning outcomes into new supervision signals, through models including rewards modeling and instruction-following task. 
Our approach is related in that it uses the model’s own generated responses (along with confidence evaluations) to create new training examples. However, we explicitly address the factuality–informativeness trade-off, which sets our work apart from general hindsight or reinforcement learning approaches that optimize for broad objectives like alignment or cumulative reward. In contrast to hindsight relabeling methods aimed at aligning with human preferences, our method builds direct supervision around factuality control, a critical dimension for high-stakes and knowledge-centric applications.



\section{Factuality-Controlled Generation}
\newenvironment{shrinkeq}[2]
{ \bgroup
  \addtolength\abovedisplayshortskip{#1}
  \addtolength\abovedisplayskip{#1}
  \addtolength\belowdisplayshortskip{#2}
  \addtolength\belowdisplayskip{#2}}
{\egroup\ignorespacesafterend}


\label{sec: method}

We propose a task called Factuality-Controlled Generation (FCG), where a language model takes as input a (question, factuality constraint) pair and outputs a response that satisfies the requested factuality level while being as informative as possible.
In essence, FCG improves the model with a controllable “knob” for factuality: given an input question and a desired factuality score $c$, the model produces an output that contains only information it is at least $c$-confident is correct, including as many details as possible under that constraint.

\subsection{Problem Setup}  
Formally, let a language model $M$ take as input a question $x \in X$ and a factuality level $c \in[0,1]$, and produce an output $o_{x, c} \in O$ (both $x$ and $o$ are natural language text). Our goal in this task is to generate an output such that $f(o_{x, c}) \geq c$ while keeping the response as informative as possible.

Our goal is to train $M$ such that for any input $(x, c)$, the output $o_{x, c}$ satisfies $f(o_{x, c}) \geq c$ while remaining as informative as possible, where $f(\cdot)$ is the factuality score defined in Section \ref{sec:eval_metrics}.

\subsection{Evaluation Metrics}
\label{sec:eval_metrics}

We evaluate each generated response along two dimensions: factuality and informativeness, $o_{x, c}$, using FactScore \cite{min-etal-2023-factscore}. We segment each output into atomic facts and and then measure:

\textbf{Factuality.} $f(o) \in [0,1]$ is the proportion of atomic facts that are verified to be correct. Following FactScore, we determine correctness by verifying each fact against Wikipedia. We additionally measure \textit{factuality adherence} at the target level $c$ via an indicator:
\begin{equation}
\mathbf{1}\left( f\left(o_{x, c}\right) \geq c \right).
\end{equation}

\textbf{Informativeness.} Informativeness is measured as the total number of atomic facts in the output, whether correct or not.




\begin{figure*}[ht!]
    \centering
    \includegraphics[width=\textwidth]{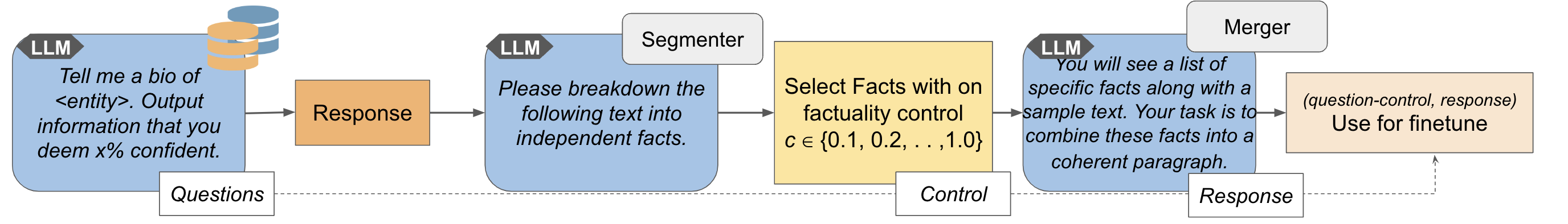}
    \caption{Synthetic data generation pipeline for creating (question, control level, response) pairs. Given a question, we generate an initial response, segment it into facts, score each fact’s confidence, and remove the lowest-confidence facts until the response’s factuality meets the control level $c$. In the prompts, we express $c $as a percentage ($c = x\%$). This yields a filtered response that adheres to the factuality constraint while maximizing informativeness, which is then used for factuality-controlled training (fine-tuning the model).
    }
    \label{fig:model}
\end{figure*}

\subsection{FCG Training}
We find that frontier models do not reliably satisfy factuality constraints via prompting alone (Section \ref{sec:training}, Table \ref{tab:results}, and Figure \ref{fig:factuality-controlled}). We therefore propose a training-based method, Factuality-Controlled Generation (FCG), which fine-tunes a language model on \textit{(question, control, response)} triples.

\subsubsection{Synthetic Data Generation}
\label{sec:data_generation}

Since our factuality-controlled setting is novel, there is no existing training corpus of \textit{(question, desired-factuality, response)} examples to use directly. Therefore, we propose an algorithm to construct a synthetic dataset $T=\{(x, c, r)\}_j^n$ where each response $r$ is intended to satisfy $f(r) \geq c$ while retaining as much information as possible.


The key idea is to leverage the model (or another LLM) to produce an initial answer and then systematically remove low-confidence facts to meet different factuality levels. Prior work has found that naively fine-tuning LLMs on only ground-truth information can sometimes reduce factuality in open-ended generation by discouraging models from providing additional details \cite{gekhman-etal-2024-fine}. In contrast, our method fine-tunes the model using responses that closely resemble its own style of output, but with minimal edits to enforce factuality constraints, thereby teaching the model how to trade off detail for accuracy in a way that aligns with its natural generation tendencies.

Specifically, for a given question $x$ and target control value $c$, we generate a response as follows. First, we prompt the base model $M$ to generate an initial response $r_0$ without any factuality conditioning: $r_0=M(x)$. If this initial response $r_0$ meets the factuality requirement (i.e., $f\left(r_0\right) \geq c$ ), we can use $\left(x, c, r_0\right)$ as a training example directly and add them to $T$. If $r_0$ fails to meet the requirement, we edit $r_0$ by progressively removing content until the factuality constraint is satisfied (meets control $c$). We remove the lowest-confidence (as determined by model $M$ ) factual statements first, under the assumption that those are most likely to be incorrect. 
This process yields a filtered response $r$ that has higher factuality (at least $c$) while retaining as much of the original information as possible. By repeating this procedure for multiple values of $c$, we obtain a set of training pairs for the question $x$ at varying factuality levels. This editing process aligns with how humans sequence information from most to least confident when responding to questions.

Formally, let $A$ be the set of all atomic facts (indivisible factual statements) that appear in a response. A segmenter $S: O \rightarrow \mathcal{P}(A)$ divides a response into atomic facts, while a merger $G: \mathcal{P}(A) \rightarrow O$ recombines a set of facts into a fluent textual response (following the approach of \cite{min-etal-2023-factscore}). Given a question–response pair with control $\left(x, c, r_0=M(x)\right)$ :

\begin{enumerate}
  \item Segment $r_0$ into atomic facts: 
        {\small
        \begin{equation}
        \left(r_0\right)=a_1, \cdots, a_n
        \end{equation}
        }
  \item Prompt model $M$ for its confidence for each fact: Estimate confidence for each fact using model $M$: for each $a_i$ in the set, query $M$ to obtain a confidence score 
        {\small
        \begin{equation}
        h_M\left(a_i\right) \in[0,1]
        \end{equation}
        }
    indicating how likely $M$ believes ai is true. In practice, we prompt the model with the fact and a question like “Is this statement True or False?” and derive a probability from its response, as detailed in Section 4.2.
  \item Rank facts by confidence in ascending order:
          {\small
        \begin{equation}
        S^{\prime}\left(r_0\right)=\left[a_1^{\prime}, \cdots, a_n^{\prime}\right], 
        \end{equation}
        }
        where $h_M\left(a_1\right) \leq h_M\left(a_2\right) \leq … \leq h_M\left(a_n\right)$.
  \item Progressively remove lowest confidence facts by traversing the sorted list and drop the lowest-confidence facts one by one until the factuality of the remaining set meets the target level $c$.  Formally, find the smallest index $j$ such that
        {\small
        \begin{equation}
        \min _{j \in\{0, \cdots, n\}} f\left(G\left(\left[a_{j+1}^{\prime}, \cdots, a_n^{\prime}\right]\right)\right) \geq c
        \end{equation}
        }
    Here, $G$ means we merge the high-confidence subset of facts back into a coherent response. By construction, this filtered response contains only facts with confidence $\geq h_M\left(a_{j+1}\right)$, which should raise its factuality.
  \item Finalize the example. Let
        {\small
        \begin{equation}
        \left(x, c, r=G\left(\left[a_{j+1}^{\prime}, \cdots, a_n^{\prime}\right]\right)\right)
        \end{equation}
        }
    be the merged response after removing the j lowest-confidence facts. Add filtered $(x, c, r)$ response to the training set $T$. If no subset of facts can meet the factuality control (e.g., $r_0$ was entirely below the threshold $c$), then we do not create an example for that $(x, c)$.
\end{enumerate}
Using the above algorithm, we generate training triples for a range of factuality levels c. In our implementation, we targeted $c \in \{0.1, 0.2, …, 1.0\}$. Each question yields up to 10 responses, from very lenient ($c=0.1$, almost no filtering) to very strict ($c=1.0$, only completely certain facts). These synthetic examples form the basis for supervised training.

\subsubsection{Fine-tuning}
After constructing the dataset $T$ of factuality-controlled examples, we perform full-parameter fine-tuning with early stopping on the synthetic (x, c, r) triples, conditioning the model on both the question x and the target factuality level c. Additional fine-tuning details shown in Section \ref{sec:training}. 
The outcome of training is a model that has learned to internalize the factuality constraint. At inference time, it can adjust its output based on the given $c$ without requiring iterative filtering.

\section{Experimental Setup}
\subsection{Dataset and Task Setup}

 \paragraph{Task.} We evaluate our method on the biography generation task as outlined by \cite{min-etal-2023-factscore}. This task involves generating biographical content for various entities, with the factual accuracy of each statement in the biography is verified against a trusted source. This task is suitable for our study because it requires a mix of factual knowledge and informative content, and it comes with an automatic metric for factuality, FactScore \cite{min-etal-2023-factscore}. FactScore assesses the precision of generated biographies by decomposing them into atomic facts and verifying each against Wikipedia articles as references, yielding a factuality score (the proportion of facts that are correct) and an informativeness measure (the total number of facts). 



\paragraph{Prompting.} To generate prompts for this task, we follow \citep{min-etal-2023-factscore} and use a simple instruction: 
\textit{“Tell me a bio of <entity>.”}. 
For incorporating factuality control, we append an additional directive such as:
\textit{“Output information that you deem <x\%> confident.”} Here, X\% is a placeholder that we replace with the desired confidence level (e.g., 80\%, 90\%, 100\%). This way, during inference, we can instruct either the base model or the fine-tuned model to attempt different factuality levels by changing the percentage in the prompt.


\paragraph{Dataset statistics.} Our training data consists of 500 entities for which we generate synthetic factuality-controlled responses as described in Section \ref{sec:data_generation}. We split these entities into 450 for training and 50 for development/validation. An additional 183 distinct entities are held out for testing. In total, the generated training set contains 3,302 (question, control, response) examples and the development set contains 396 examples, each question being paired with multiple responses at various factuality levels in the form of $\left(x, c, r\right)$ triples. All experiments are conducted on these biography data splits.

\paragraph{Evaluation}
We use FactScore \cite{min-etal-2023-factscore} for automated factuality evaluation, and follow the metrics defined in Section \ref{sec:eval_metrics}.

\subsection{Synthetic Data Generation}
To generate \textit{(question-control, response)} data needed for fine-tuning, we use GPT-4 as both segmenter $S$ and merger $G$, as described in Section \ref{sec:data_generation}. The segmenter $S$ takes a model-generated response and breaks this into individual facts, prompted by:

\textit{"Please breakdown the following text into independent facts."}

This yields a list of atomic factual statements. 

The merger $G$ is given a subset of facts as input, which includes selected facts $\left[a_{j+1}^{\prime}, \ldots, a_n^{\prime}\right]$, along with the original response $r_0$. It merges these facts while preserving $r_0$ 's phrasing where possible, to maintain similarity, as detailed in Section \ref{sec: method}. The merger's prompt is:

\textit{"You will see a list of specific facts along with a sample text. Your task is to combine these facts into a coherent paragraph. If a fact from the list is also in the sample text, use the same phrasing as the sample text. Ensure that you exclude any details not listed in the facts, even if they appear in the sample text."}

This procedure ensures that the filtered response remains fluent and as close as possible to the original phrasing, changing only by the removal of low-confidence parts.

To assess model $M$ 's confidence in atomic fact $a_i$, we use the prompt:

\textit{
"<fact> Is this statement True or False? Start your answer with either "True" or "False"."}


We then evaluate $M$’s next-token probabilities for the words “True” vs “False”. The normalized probability assigned to “True” is taken as the confidence score $h_M(a_i)$ for that fact. Essentially, this uses the model’s internal judgment as a numeric confidence estimate via a language-model-based binary classifier. This method provides a relative measure of the model’s belief in the fact’s correctness.

\begin{table}[t!]
\centering
\resizebox{\columnwidth}{!}{%
\begin{tabular}{lccc}
\hline
                                & $c=0.8$ & $c=0.9$ & $c=1.0$ \\ \hline
No Factuality Control           & 15.9    & 5.5     & 0.0     \\
Factuality-Controlled Inference & 14.8    & 3.8     & 0.0     \\
Factuality-Controlled Training & $\mathbf{1 8 . 7}$ & $\mathbf{1 2 . 6}$ & $\mathbf{2 3 . 6}$ \\ \hline
\end{tabular}%
}
\caption{ Factuality adherence (percentage of outputs satisfying the factuality constraint $c$) for different methods on the test set. Our Factuality-Controlled Training significantly improves factuality satisfaction across high confidence levels. }
\label{tab:results}
\end{table}

\subsection{Model and Training Details}
\label{sec:training}
As language model $M$ for our experiments, we use Mistral-7B-Instruct-v0.2 \cite{jiang2023mistral}, a 7-billion-parameter transformer model fine-tuned on instruction-following tasks. We then perform Factuality-Controlled Generation fine-tuning on Mistral-7B. 

We fine-tune all model parameters with early stopping based on validation loss using the generated data. We evaluate learning rates of 3e-6, 1e-5, 3e-5, 1e-4 to identify the optimal configuration. 
Fine-tuning is conducted for up to 30 epochs, with early stopping triggered after 4 consecutive epochs of no improvement in development set loss. 
The batch size is set to 256.

\subsection{Baselines}
We compare our Factuality-Controlled Generation fine-tuning method to two baselines:

1. No factuality control (NFC): The prompts contain only the questions prompt (e.g., “Tell me a bio of X.”) without any specification of factuality. This represents the vanilla generation behavior without any control mechanism.

2. Factuality-controlled inference (FCI): the model is $not$ fine-tuned on factuality-controlled data, but at test time we prepend the same factuality instructions. For example, “Output information you deem 90\% confident”. This is the similar approach that inspired our training data, applied directly as an inference-time control.

These baselines allow us to compare the improvements of our fine-tuning. FCI represents the straightforward application of our content removal strategy without additional training, while NFC shows the model’s unconstrained output. We evaluate whether FCG fine-tuning yields improvements over simply prompting or filtering the base model. Table \ref{tab:baseline-summary} summarizes the differences between the methods.

\begin{table}[t]
\centering
\small
\setlength{\tabcolsep}{3.5pt}
\renewcommand{\arraystretch}{1.05}
\resizebox{\columnwidth}{!}{%
\begin{tabular}{lcc}
\toprule
Method & with FCG prompting? & with FCG training? \\
\midrule
No factuality control (NFC) & No & No \\
Factuality-controlled inference (FCI) & Yes & No \\
Factuality-controlled training (FCG) & Yes & Yes \\
\bottomrule
\end{tabular}%
}
\caption{Summary of the differences between the three methods compared in our experiments.}
\label{tab:baseline-summary}
\end{table}


\section{Results}

\begin{figure}[t]
    \centering
    \includegraphics[width=1.0\linewidth]{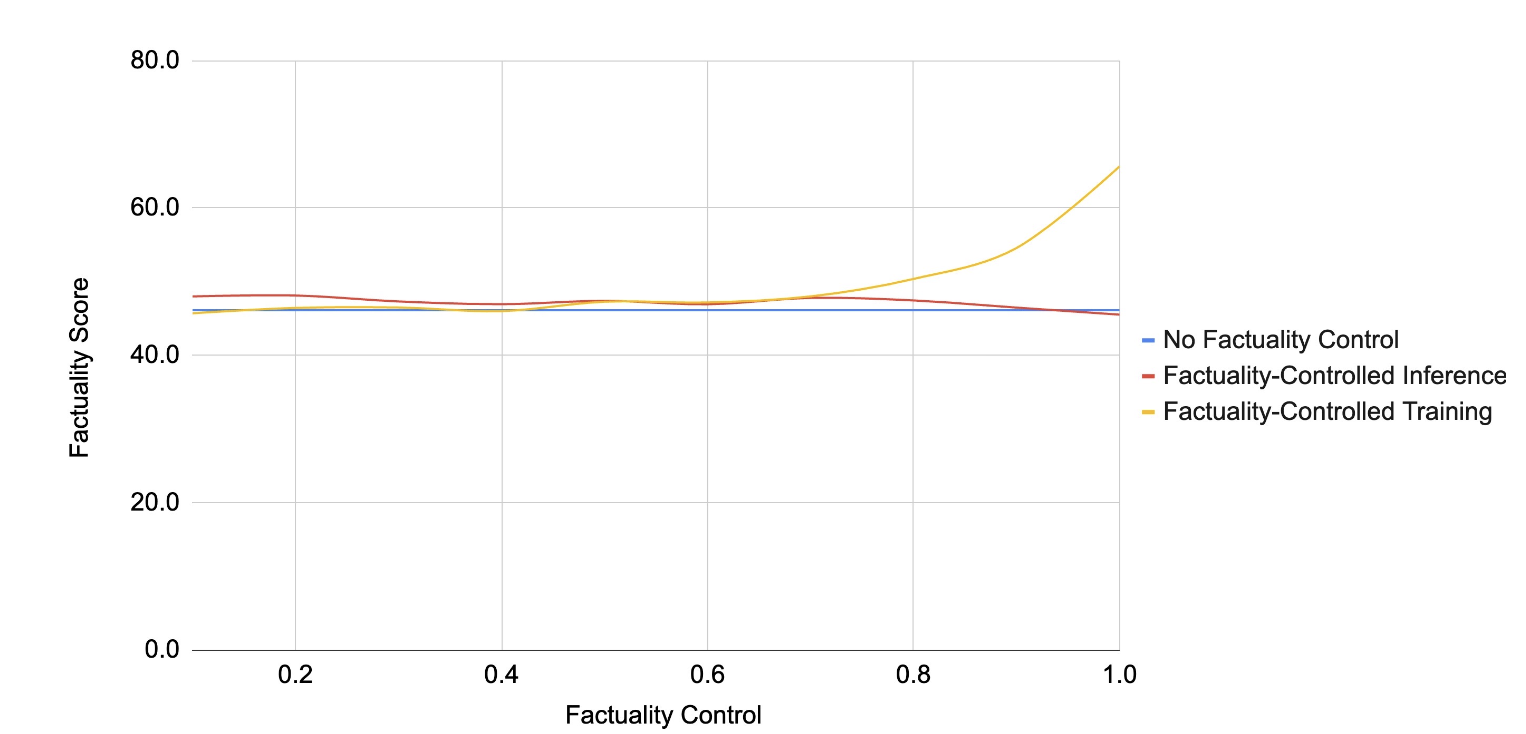}
    \caption{
    Our factuality-controlled training method improves the model's ability to respect factuality constraints: as the target factuality level $c$ increases, the model's outputs become more factual (yellow line). In contrast, the baseline models show no consistent improvement (red and blue lines).
    }
    \label{fig:factuality-controlled}
\end{figure}

We evaluate the proposed factuality-controlled training method and the baselines along two complementary dimensions. First, we measure factuality adherence: how often a model respects the requested factuality constraint $c$ (Section \ref{sec:eval_metrics}; Table \ref{tab:results} and Figure \ref{fig:factuality-controlled}). Second, we analyze the factuality--informativeness trade-off, i.e., how much information each method provides at a given factuality level (Figure \ref{fig:informativeness}).


\subsection{Factuality Adherence} 
\textbf{Factuality-controlled training significantly improves factuality satisfaction.} Table \ref{tab:results} reports the factuality satisfaction rates for outputs at different target levels $c$. 
Notably, the FCI baseline (prompting the base model with confidence instructions, without fine-tuning) does not reliably improve factuality over the no-control case. In fact, we observe that FCI slightly underperforms the vanilla model on some settings. For example, at c = 0.9, only 3.8\% of FCI outputs met the threshold, versus 5.5\% with no control. This counterintuitive result suggests that the base Mistral-7B model struggles to follow the raw factuality prompt; simply instructing it to “be more confident” does not yield more factual answers, and the added instruction might even confuse the generation process, leading to fewer correct facts. In contrast, our fine-tuned model (FCG) shows substantially better adherence: 18.7\% of its outputs meet c = 0.8 (a 17.2\% relative improvement over 15.9\%), 12.6\% meet c = 0.9 (a 130\% relative improvement over 5.5\%), and 23.6\% meet the strict c = 1.0 criterion (where neither baseline model produced any fully factual output). These improvements indicate that the ability to control factuality can indeed be instilled via supervised training. The FCG model has learned to adjust its content and only include facts that it is sufficiently confident in, whereas the off-the-shelf model could not effectively utilize the control signal on its own.

\textbf{FCG learns to use the control signal.} To validate that the model learned the intended behavior, we plot the average factuality of outputs vs. the requested level in Figure \ref{fig:factuality-controlled}. Before fine-tuning, the base model’s factuality does not consistently increase when higher confidence levels are requested. After fine-tuning, there is a clear upward trend: higher target factuality $c$ yields higher actual factuality in the outputs. This indicates that FCG training teaches the model to produce increasingly factual outputs as the target level increases.


\subsection{Factuality–Informativeness Trade-off}
\begin{figure}[t]
    \centering
    \includegraphics[width=\linewidth]{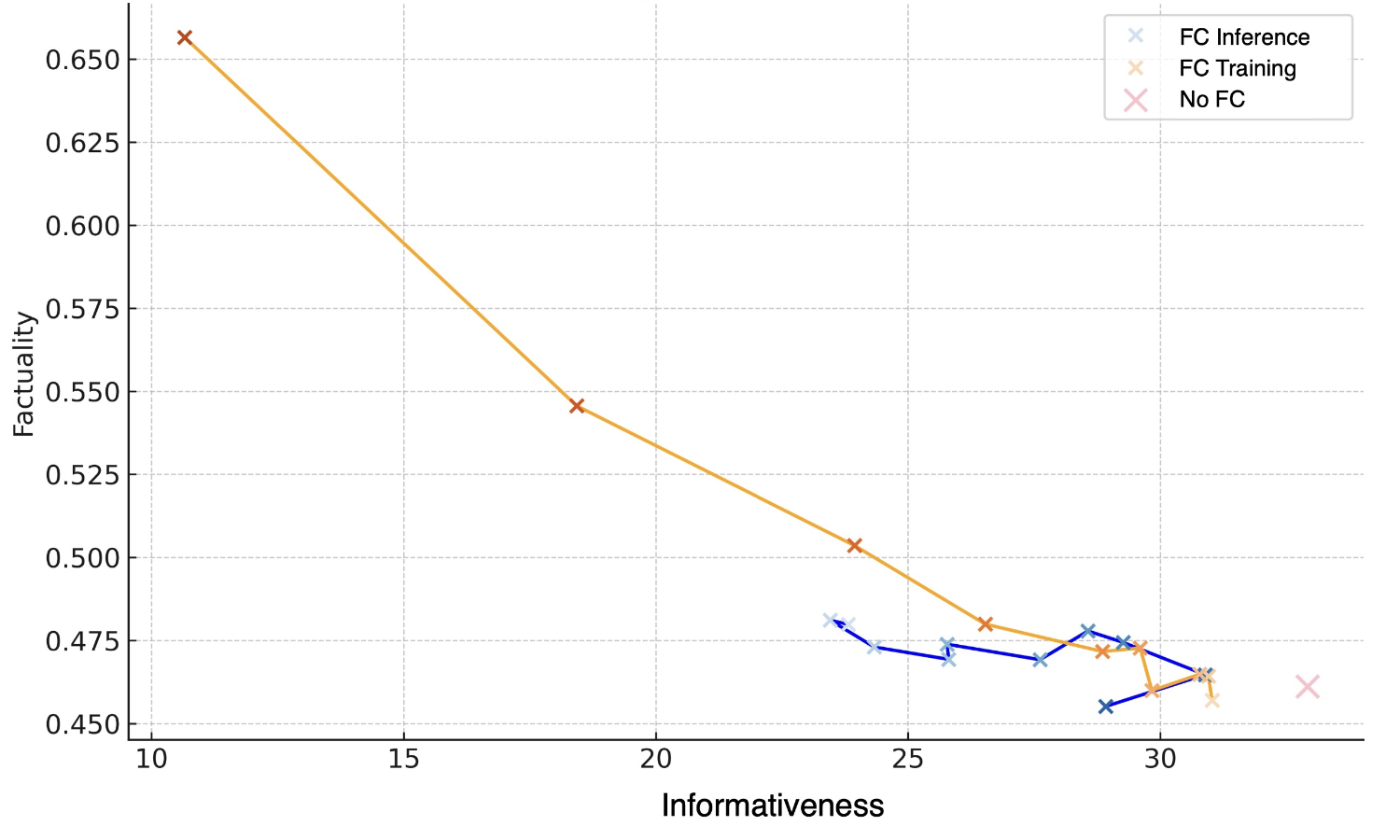}
    \caption{ Factuality vs. informativeness trade-off curves for three methods. No factuality control (NFC) uses the base model with no control instruction. Factuality-controlled inference (FCI) uses the base model with a test-time control instruction but no fine-tuning. Factuality-controlled training (FCG) fine-tunes the model on synthetic (question, control, response) data and applies the same test-time control. FCG achieves a better trade-off than the baselines (higher informativeness at the same factuality), and at $c = 1.0$ it still produces informative responses (23.6\% of outputs contain only verified facts), whereas the baselines produce no fully factual outputs.
    }
    \label{fig:informativeness}
\end{figure}

\textbf{FCG improves the trade-off between factuality and informativeness.}
Figure \ref{fig:informativeness} plots the factuality vs. informativeness curves achieved by the different methods (varying $c$ yields different points on each model’s curve). An ideal method would lie toward the top-right (high factuality, high informativeness). 
Our fine-tuned model (FCG) generally dominates the baselines. In particular, it outperforms the FCI at most operating points. The FCG curve is on the outer frontier, meaning that for the same factuality level, FCG produces more informative responses. For example, at a factuality of roughly 0.9, FCG outputs contain more facts on average than FCI outputs at a comparable factuality. The gap between FCG and FCI is not large (the curves are close, reflecting that our method and the baseline ultimately use a similar filtering principle), but FCG is never worse than FCI and has a clear advantage in the high-factuality regime. Notably, at the strictest setting (c = 1.0), FCG can still provide some information (informativeness > 0), whereas the inference baseline effectively must prune almost everything (the base model always included at least one low-confidence fact, leading FCI to drop all content, hence 0 informativeness for a truly 100\% factual output). This highlights the benefit of training: the fine-tuned model can find ways to rephrase or include only guaranteed facts, yielding a non-empty answer that is fully factual – something the baseline model did not achieve.

While our method explicitly controls the percentage of correct facts in a response, informativeness is implicitly optimized by allowing the model to generate as much factual content as possible while adhering to the specified factuality constraint. Thus, the absolute informativeness is bounded by what the base model can produce. Thus, when a high factuality constraint is in place, the model prioritizes factually verifiable statements while still including as much relevant information as possible. Conversely, the model has the freedom to incorporate a broader range of details, including some that are less verifiable or more speculative, resulting in higher informativeness (more facts mentioned) at the cost of some accuracy. This behavior aligns with our design of the training data: because we always removed the minimum necessary facts, the model learned “if you must be x\% factual, drop the least-certain details but keep everything else.” The end result is that FCG improves the factuality–informativeness trade-off: for any given level of factual accuracy, it attempts to make the response as informative as possible within that constraint.

\textbf{Synthetic training shifts the trade-off curve outward.} Because our training examples are constructed by removing the minimum amount of low-confidence content needed to satisfy each factuality constraint, the model learns to keep higher-confidence facts and drop lower-confidence details as $c$ increases. As a result, for any target factuality level, FCG produces responses that are as informative as possible while respecting the constraint.

\section{Conclusion} 
We introduced Factuality-Controlled Generation (FCG), a framework and evaluation setup for conditioning LLM outputs on a user-specified factuality level and measuring the resulting factuality--informativeness trade-off.
We also showed that a simple instantiation of FCG based on fine-tuning with synthetic data can improve adherence to the requested factuality level while maintaining informativeness.
Our experiments show that \emph{i)} FCG significantly improves a model’s ability to satisfy factuality requirements. \emph{ii)} FCG also enhances the factuality–informativeness trade-off across evaluated settings, achieving comparable or greater informativeness than baseline methods at the same factuality level. These results validate FCG as an effective method to improving the controllability of text generation, allowing LLMs to be more reliably adapted to different needs.

\textbf{For future work}, we hope our methods and findings will inspire further research on controllable text generation, including \emph{i)} scaling FCG training to larger models to evaluate its performance on state-of-the-art architectures, \emph{ii)} applying FCG to more complex tasks, such as domain-specific generation or multimodal understanding, \emph{iii)} investigating how factuality and informativeness constraints are represented within a model’s parameters,  \emph{iv)} exploring alternative methods for improving controllability and flexibility in language generation, and \emph{iv)} incorporating more diverse editing operations in the synthetic data generation process, which could further improve the balance between factual accuracy and informativeness. We believe addressing these questions will help develop LLMs that are both reliable and flexible, adapting their output to the specific factual requirements of any given application.

    
\section*{Limitations}
While our work introduces a novel framework for Factuality-Controlled Generation and demonstrates its effectiveness, several limitations should be acknowledged: First, we focus on biography generation and evaluate factuality using FactScore. This limited scope does not cover other generative tasks. Future work should explore diverse applications, such as abstractive summarization and domain-specific tasks, to assess generalizability. Second, this study is limited to English. While the FCG framework is theoretically applicable to other languages, its effectiveness remains unexplored. Extending FCG to multilingual settings is an important direction for future work. Third, our experiments are conducted on a mid-sized model, and the scalability of FCG to larger state-of-the-art models remains untested. Evaluating its performance and effectiveness at scale is essential for broader adoption.

\section*{Acknowledgments}
 This work was in part supported by the funds provided by the National Science Foundation and by DoD OUSD (R\&E) under Cooperative Agreement PHY-2229929 (The NSF AI Institute for Artificial and Natural Intelligence). The views, opinions and/or findings expressed are those of the authors and should not be interpreted as representing the official views or policies of the National Science Foundation or the U.S. Government.

\bibliography{anthology,custom}

\appendix



\end{document}